\documentclass{article}

\usepackage{arxiv}

\usepackage{kotex}
\usepackage{hyperref}       
\usepackage{url}            
\usepackage{booktabs}       
\usepackage{amsfonts}       
\usepackage{nicefrac}       
\usepackage{microtype}      
\usepackage{lipsum}
\usepackage{graphicx}
\usepackage{array}
\graphicspath{ {./images/} }
\usepackage{hyperref}
\usepackage{tabularx}
\usepackage{multirow}
\usepackage{float}

\title{KoACD: The First Korean Adolescent Dataset for Cognitive Distortion
Analysis via Role-Switching Multi-LLM Negotiation}

\author{
 JunSeo Kim \\
  Department of Computer Engineering\\
  Gachon University College of IT Convergence\\
  Seongnam, 13120 \\
  \texttt{kma80kjs@gachon.ac.kr} \\
   \And
 HyeHyeon Kim \\
  Department of Biomedical Systems Informatics\\
  Yonsei University College of Medicine\\
  Seoul, 03722 \\
  \texttt{hye\_hyeon\string@yonsei.ac.kr}
}

\begin{document}
\maketitle
\begin{abstract}
Cognitive distortion refers to negative thinking patterns that can lead to mental health issues like depression and anxiety in adolescents. Previous studies using natural language processing (NLP) have focused mainly on small-scale adult datasets, with limited research on adolescents. This study introduces KoACD, the first large-scale dataset of cognitive distortions in Korean adolescents, containing 108,717 instances. We applied a multi-Large Language Model (LLM) negotiation method to refine distortion classification, enabling iterative feedback and role-switching between models to reduce bias and improve label consistency. In addition, we generated synthetic data using two approaches: cognitive clarification for textual clarity and cognitive balancing for diverse distortion representation. Validation through LLMs and expert evaluations showed that while LLMs classified distortions with explicit markers, they struggled with context-dependent reasoning, where human evaluators demonstrated higher accuracy. KoACD aims to enhance future research on cognitive distortion detection. The dataset and implementation details are publicly accessible.\footnote{\href{https://github.com/cocoboldongle/KoACD}{https://github.com/cocoboldongle/KoACD}}
\end{abstract}


\section{Introduction}

Negative thoughts \cite{pietromonaco1985nature} are a natural part of human cognition, often helping individuals recognize potential dangers, prepare for challenges, or engage in self-reflection. However, when these patterns become rigid and excessive, they can lead to emotional distress and contribute to mental health issues such as depression and anxiety disorders. Adolescents, in particular, may be more vulnerable to these maladaptive thought patterns due to their ongoing cognitive and emotional development.

Globally, one in seven children and adolescents (about 166 million) suffers from mental illness, with 42.9\% experiencing anxiety and depression \cite{unicef2018}. The rising prevalence of these conditions during adolescence has become a serious global concern. Since this stage is crucial for self-identity formation and emotional regulation \cite{pfeifer2018development}, understanding how negative thought patterns emerge and persist is essential for early intervention and prevention.

In particular, depression is frequently associated with habitual negative thinking, known as cognitive distortion \cite{rnic2016cognitive}. Adolescents experiencing these distortions may instinctively blame themselves when something goes wrong, thinking, ``I messed up again'' or ``I completely failed.'' These persistent negative thoughts trigger emotional distress, reinforcing cycles of depression \cite{mahali2020associations}. Identifying and analyzing these patterns is essential for developing effective coping strategies. To achieve this, building a comprehensive dataset of adolescent cognitive distortions is necessary to enable more targeted and impactful mental health interventions.

Our key contributions are as follows:
\begin{itemize}
    \item We present KoACD, the first large-scale dataset of cognitive distortions in Korean adolescents, filling a critical gap in non-English and youth-oriented mental health resources.
    \item We propose a multi-round negotiation framework using multiple Large Language Models (LLMs) to refine cognitive distortion classification, employing iterative role-switching to improve consistency, reduce bias, and capture context-dependent reasoning.
\end{itemize}

\section{Related Work}

\subsection{Cognitive Distortions Detection}

Cognitive distortions are closely related to negative thinking patterns and are defined as distorted ways of thinking that reinforce negative emotions \cite{beck1979cognitive}. With the recent development of natural language processing (NLP), research has been actively conducted to automatically classify cognitive distortions using various datasets.

Previous studies on cognitive distortion classification have primarily relied on small-scale, adult-focused, and English-language datasets. Early research utilized Linguistic Inquiry and Word Count (LIWC)-based regression models on social media posts \cite{simms2017detecting}, while later studies adopted deep learning models, including RNNs, CNNs, and BERT, using datasets from counseling platforms and therapist-patient conversations \cite{shickel2020automatic, tauscher2023automated}. More recent approaches have leveraged Large Language Models (LLMs) for cognitive distortion classification, further enhancing performance and adaptability across diverse datasets \cite{chen2023empowering, qi2023supervised}.

Table~\ref{tab:dataset_summary} summarizes existing datasets that deal with cognitive distortions.

Despite these advancements, existing datasets remain limited in scale and predominantly focus on English-speaking adults. To address these limitations, we propose a Korean-language dataset specifically designed for adolescents, filling a crucial gap in research on cognitive distortions in younger populations.

\begin{table}[htbp]
\centering
\renewcommand{\arraystretch}{1.25}
\setlength{\tabcolsep}{6pt}
\begin{tabular}{>{\raggedright\arraybackslash}p{4cm}lllll}
\toprule
\textbf{Dataset} & \textbf{Language} & \textbf{Sample} & \textbf{Target} & \textbf{Data Source} & \textbf{Classification} \\
\midrule
Tumblr\\Cognitive Distortion* \cite{simms2017detecting} & English & 459 & nonspecific & Tumblr blogs & Binary (2) \\
MH-C \cite{shickel2020automatic} & English & 1,164 & Adult & TAO Connect & Multi-class (15) \\
MH-D \cite{shickel2020automatic} & English & 1,799 & Adult & TAO Connect & Binary (2) \\
CrowdDist \cite{shickel2020automatic} & English & 7,666 & Adult & Mechanical Turk & Multi-class (15) \\
Clinician-Client\\SMS* \cite{tauscher2023automated} & English & 7,354 & Adult & Clinician-Client SMS & Multi-class (5) \\
SocialCD-3K \cite{qi2023supervised} & Chinese & 3,407 & nonspecific & Weibo & Multi-label (12) \\
\bottomrule
\end{tabular}

\setlength{\abovecaptionskip}{10pt} 
\caption{Summary of datasets for cognitive distortion detection. The 'Sample' column indicates the number of instances, and 'Classification' specifies the type (binary, multi-class, or multi-label). *Indicates unofficially named datasets.}
\label{tab:dataset_summary}
\end{table}

\subsection{LLMs-Based Negotiation}

Attempts have been made to explore the possibility of models going beyond independent judgment through interaction and negotiation between LLMs.

Self-Play and In-Context Learning techniques using AI feedback were applied to improve the negotiation capabilities of LLMs \cite{fu2023improving}, and a method for providing feedback by developing an LLM-based Assistant for Coaching nEgotiation (ACE) using negotiation data from MBA students was proposed \cite{shea2024ace}. In addition, research has been conducted on applying negotiation methods to emotional analysis. It has been demonstrated that using LLM negotiation methods to interact between models can outperform the existing single-pass decision \cite{sun2023sentiment}.

Previous studies have shown that LLM negotiations can yield sophisticated results, but challenges remain in balancing negotiation outcomes due to fixed roles and limited structures. To overcome this, we use role-switching and multiple types of LLMs to balance the negotiation process. We also introduce multi-round negotiations to give equal consideration to the 10 cognitive distortions, ultimately arriving at the optimal conclusion. Therefore, this study aims to generate and validate data using LLM negotiation techniques.

\section{Constructing KoACD}

\subsection{Data Source and Preprocessing}

We crawled posts on NAVER Knowledge iN\footnote{\url{https://kin.naver.com/}}, a Q\&A platform where users can post questions and receive answers, to analyze the cognitive distortions of Korean adolescents. NAVER Knowledge iN is Korea’s largest open Q\&A platform, with over 32 million users and 800 million Q\&A entries since its launch \cite{jang2024key}. The use of NAVER Knowledge iN complies with its terms of service and copyright regulations and is permitted through prior approval or explicit permission. Since NAVER Knowledge iN covers a wide range of age groups, we used only data from five major adolescent counseling organizations and services, covering the years 2011--2024, to focus on adolescent concerns. A total of 69,925 questions were collected, and the distribution of data sources is provided Appendix~\ref{appendix:data-distribution}.

A pre-processing step refined the data to align with the research purpose and excluded irrelevant questions. This included removing entries from elementary students or adults, filtering inappropriate content, deleting vague questions, and eliminating duplicates. After applying these criteria, 37,124 questions remained for analysis. Details on preprocessing and removed cases are in Appendix~\ref{appendixB}.

\subsection{Definition of Cognitive Distortions}

Aaron Beck, a pioneer in cognitive therapy, identified 10 cognitive distortions in patients with depression and incorporated them into psychotherapy \cite{beck1979cognitive}. He emphasized that reducing these distortions could alleviate stress and anxiety \cite{beck1991cognitive}. We used these distortions, listed in Table~\ref{tab:cognitive_distortions}, to classify questions reflecting the emotional struggles commonly reported by adolescents.

\begin{table}[htbp]
\centering
\renewcommand{\arraystretch}{1.3}
\setlength{\tabcolsep}{6pt}
\begin{tabular}{
>{\raggedright\arraybackslash}p{4.5cm} 
>{\raggedright\arraybackslash}p{6.5cm} 
>{\raggedright\arraybackslash}p{4.0cm}
}
\toprule
\textbf{Cognitive Distortion Type} & \textbf{Definition} & \textbf{Examples} \\
\midrule
All-or-Nothing Thinking & Viewing situations in only two categories (e.g., perfect or failure) instead of on a spectrum. & "If I fail this test, I'm a total failure." \\
Overgeneralization & Drawing broad conclusions from a single event or limited evidence. & "My one friend ignored me, so everyone else will hate me too." \\
Mental Filtering & Focusing only on the negative aspects of a situation while ignoring the positive. & "I only remember my mistake though I got compliments on my presentation." \\
Discounting the Positive & Rejecting positive experiences or compliments by insisting they don’t count. & "People told me I did well, but I was just being polite." \\
Jumping to Conclusions & Predicting negative outcomes without evidence. & "She didn’t text back. She must be mad at me." \\
Magnification and Minimization & Exaggerating negative or risky aspects while minimizing positive aspects. & "One little mistake at work means I'm incompetent." \\
Emotional Reasoning & Believing something must be true because you feel it strongly. & "I feel worthless, so I must be worthless." \\
``Should'' Statements & Holding rigid rules about how you or others should behave, leading to guilt or frustration. & "I should always be productive; otherwise, I'm lazy." \\
Labeling & Assigning negative labels to yourself or others based on one event. & "I made a mistake, so I'm a total failure." \\
Personalization & Blaming yourself for events outside your control or assuming excessive responsibility. & "My friend looks sad, maybe I did something wrong." \\
\bottomrule
\end{tabular}

\setlength{\abovecaptionskip}{10pt} 
\caption{Classification of cognitive distortions with definitions and examples.}
\label{tab:cognitive_distortions}
\end{table}

\subsection{Multi-LLMs Negotiation for Identifying Cognitive Distortions}

To effectively identify cognitive distortions, this study designed a process for deriving optimal distortions using the multi-LLM negotiation method, where relevant distortions are gradually derived through LLM interactions \cite{fu2023improving}.

This study uses a multi-LLM negotiation method based on the interaction between Google's Gemini 1.5 Flash \cite{gemini2024} and OpenAI's GPT-4o mini \cite{openai2024gpt4omini}. The two models work together to identify the most accurate cognitive distortion. One model acts as the Analyzer and the other as the Evaluator. Through their collaboration, cognitive distortions are gradually refined. Negotiation is conducted up to five rounds to systematically explore all ten predefined cognitive distortions, as each round consists of two turns, evaluating one distortion per turn. This structure allows for the possibility that the same sentence may be interpreted with multiple cognitive distortions before reaching a final classification. Table~\ref{tab:llm-hyperparams} in Appendix~\ref{appendixE} details the LLMs parameters used in this process.

Here’s how the roles work:
\begin{itemize}
    \item \textbf{Analyzer}: Identifies the most relevant cognitive distortion in a sentence and suggests sentences that match it.
    \item \textbf{Evaluator}: Reviews the suggestions made by the Analyzer and provides feedback on their accuracy.
\end{itemize}

The prompts used for these roles are detailed in Appendix~\ref{appendixH}.
In each round of negotiation, the models take turns playing the roles of Analyzer and Evaluator.

A round consists of two turns and proceeds in the following structure:
\begin{itemize}
    \item \textbf{T1 (Initial Analysis)}: Identify the most relevant cognitive distortion in the sentence (options: one of the ten cognitive distortions).
    \item \textbf{T1 (Evaluation)}: Assess whether the proposed cognitive distortion from T1 (Initial Analysis) accurately reflects the distortion present in the sentence (options: "Yes" or "No"; the evaluator provides a justification).
    \item \textbf{T2 (Reanalysis)}: If T1 (Evaluation) results in rejection, select the next most relevant cognitive distortion, excluding previously rejected options (options: one of the remaining cognitive distortions).
    \item \textbf{T2 (Evaluation)}: Determine whether the cognitive distortion from T2 (Reanalysis) is appropriate (options: "Yes" or "No"; the evaluator provides a justification).
\end{itemize}

Each step is performed sequentially, incorporating feedback from the previous evaluation. T1 (Evaluation) assesses the distortion proposed in T1 (Initial Analysis), and T2 (Reanalysis) refines the selection based on that feedback. Similarly, T2 (Evaluation) verifies the suitability of the distortion chosen in T2 (Reanalysis).

Throughout the negotiation process, distortions deemed inappropriate are systematically excluded, ensuring the selection of the most fitting cognitive distortion. To maintain fairness, the models alternate roles in T2 (Reanalysis) so that both contribute equally to the negotiation.

If a consensus is not reached after five rounds, the question is classified as unknown. This indicates that all cognitive distortions proposed during the negotiation process were considered inherently inappropriate.

\begin{figure}[ht]
  \centering
  \includegraphics[width=0.9\linewidth]{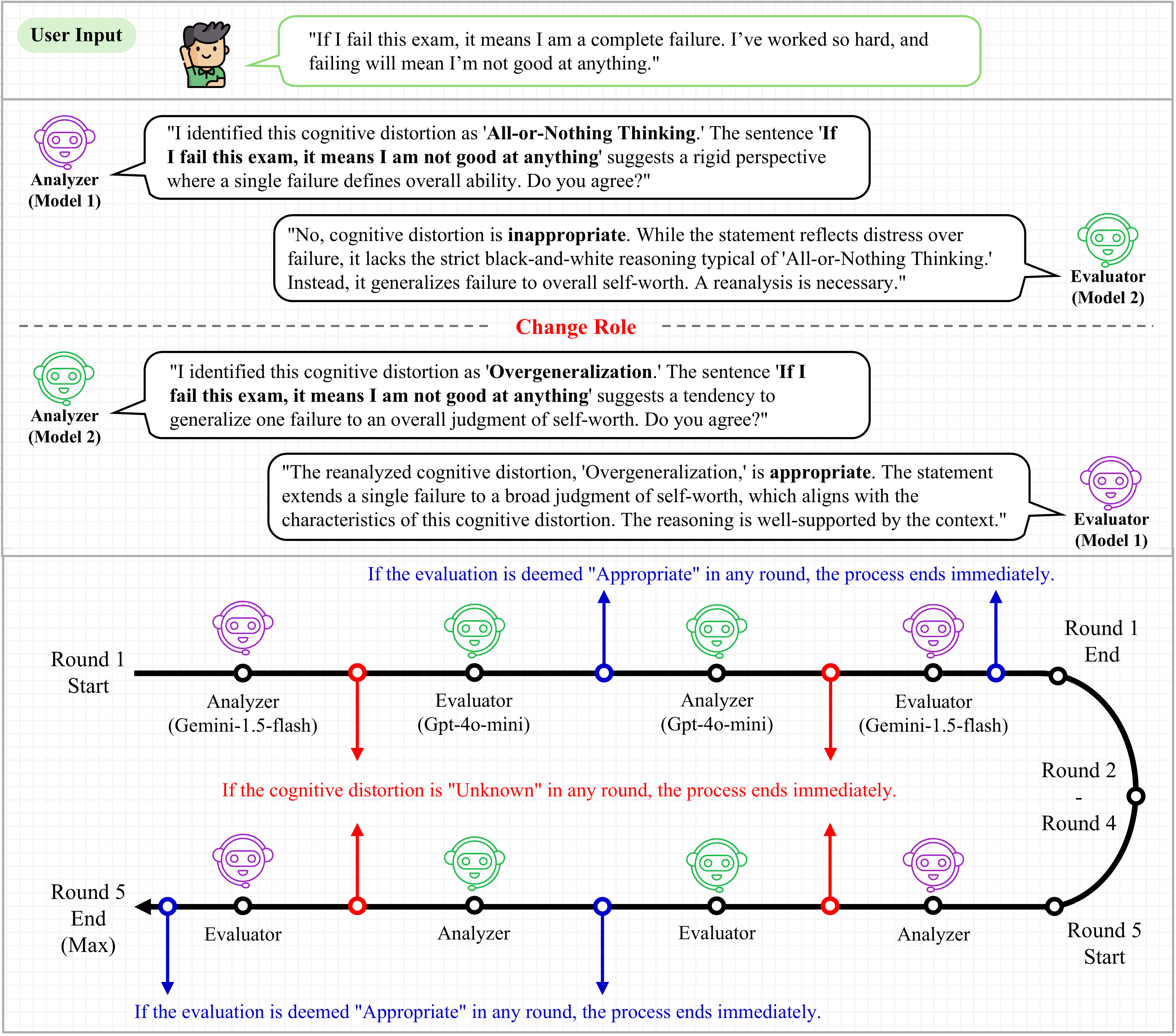}
  \caption{Process for identifying and evaluating cognitive distortions through negotiation.}
  \label{fig:negotiation-structure}
\end{figure}

The number of turns required to identify cognitive distortions or classify the question as unknown varies across datasets. Some sentences reach conclusions early, while others require multiple turns for final classification. Details of turn counts and classification ratios are given in Appendix~\ref{appendixC}. The overall structure of this negotiation process is illustrated in Figure~\ref{fig:negotiation-structure}.

\subsection{Independent Evaluation}

After the negotiation process is complete, Anthropic's Claude 3 Haiku \cite{anthropic2024claude} is used for an independent validation of the final cognitive distortion and its corresponding sentence. Claude 3 Haiku is not involved in the negotiation process; instead, it evaluates whether the selected cognitive distortion correctly aligns with the given sentence.

During negotiation, the models assess cognitive distortions in the context of the entire original text, whereas Claude 3 Haiku determines the appropriateness based solely on the selected sentence. This additional validation step helps identify potential misclassifications and ensures that the cognitive distortion is properly connected to the sentence.

Claude 3 Haiku assigns a relevance score from 1 to 3, and only cognitive distortion–sentence pairs that receive a score of 3 are used as the final-stage data for generating synthetic data. A summary of the validation score distribution is presented in Table~\ref{tab:validation_scores}. The parameters used for this validation are detailed in Table~\ref{tab:llm-hyperparams} (Appendix~\ref{appendixE}, and the prompts used for evaluation are provided in Appendix~\ref{appendixH}.

\begin{table}[ht]
\centering
\caption{Distribution of validation scores.}
\label{tab:validation_scores}
\begin{tabular}{lrr}
\toprule
\textbf{Score} & \textbf{Count} & \textbf{Proportion (\%)} \\
\midrule
1 & 11 & 0.06 \\
2 & 874 & 4.41 \\
3 & 18,897 & 95.53 \\
\midrule
Total & 19,782 & 100.00 \\
\bottomrule
\end{tabular}
\end{table}

\subsection{Synthetic Data Generation}

The original data consists of free-form text written by adolescents, often containing spelling errors, excessive use of emoticons, or unclear wording, making it difficult to interpret. Additionally, some texts lack contextual coherence, with disjointed narratives or insufficient background information to accurately assess cognitive distortions. As a result, the data could be difficult to use as is.

Furthermore, the distribution of the ten cognitive distortion categories we propose was imbalanced, leading to a potential bias in the dataset. To address these issues, we employ two methods to generate synthetic data. The prompts used for both methods are provided in Appendix~\ref{appendixH}.

\subsubsection{Cognitive Clarification of Cognitive Distortions}

The first approach to generating synthetic data is to identify cognitive distortions and rephrase the text in a clearer and more structured form while preserving the meaning of the original text, maintaining the emotional tone and context.

We used three LLMs—Gemini 1.5 Flash, Claude 3 Haiku, and GPT-4o mini—independently to generate a wide variety of expressions from 18,897 samples, ensuring greater diversity in the generated content. The parameters of these models used for synthetic data generation are detailed in Table~\ref{tab:llm-hyperparams} in Appendix~\ref{appendixE}.

\subsubsection{Balancing Cognitive Distortions with Context-Preserved Data}

The second approach we adopted aimed to address the imbalance of cognitive distortions by utilizing data classified as "Unknown" or data for which a suitable cognitive distortion could not be identified. First, we analyzed the distribution of cognitive distortions to detect which types were underrepresented. Then, synthetic data was generated by reconstructing and reorganizing 17,342 samples labeled as "Unknown," ensuring that the overall context was preserved.

Table~\ref{tab:synthetic_distribution} summarizes the distribution of cognitive distortions produced through both the cognitive clarification and cognitive balancing methods, along with the overall total after combining both approaches.

\begin{table}[htbp]
\centering
\renewcommand{\arraystretch}{1.2}
\setlength{\tabcolsep}{6pt}
\begin{tabular}{p{4.3cm} r r r}
\toprule
\textbf{Cognitive Distortion Type} & \textbf{Cognitive Clarification (\%)} & \textbf{Cognitive Balancing (\%)} & \textbf{Total (\%)} \\
\midrule
All-or-Nothing Thinking & 5,949 (10.50\,\%) & 4,920 (9.46\,\%) & 10,869 (10.00\,\%) \\
Overgeneralization & 11,418 (20.14\,\%) & 0 (0.00\,\%) & 11,418 (10.50\,\%) \\
Mental Filtering & 2,763 (4.88\,\%) & 8,139 (15.64\,\%) & 10,902 (10.03\,\%) \\
Discounting the Positive & 822 (1.45\,\%) & 9,873 (18.98\,\%) & 10,695 (9.84\,\%) \\
Jumping to Conclusions & 10,479 (18.48\,\%) & 183 (0.35\,\%) & 10,662 (9.81\,\%) \\
Magnification and \newline Minimization & 6,078 (10.72\,\%) & 4,836 (9.30\,\%) & 10,914 (10.04\,\%) \\
Emotional Reasoning & 10,842 (19.12\,\%) & 0 (0.00\,\%) & 10,842 (9.98\,\%) \\
Should Statements & 2,697 (4.76\,\%) & 7,998 (15.37\,\%) & 10,695 (9.84\,\%) \\
Labeling & 2,373 (4.19\,\%) & 8,463 (16.27\,\%) & 10,836 (9.97\,\%) \\
Personalization & 3,270 (5.77\,\%) & 7,614 (14.63\,\%) & 10,884 (10.01\,\%) \\
\midrule
Total & 56,691 (100.00\,\%) & 52,026 (100.00\,\%) & 108,717 (100.00\,\%) \\
\bottomrule
\end{tabular}

\setlength{\abovecaptionskip}{10pt} 
\caption{Distribution of cognitive distortion types across synthetic data generation methods.}
\label{tab:synthetic_distribution}
\end{table}

\section{Validating Synthetic Data with Clustering}

To verify the validity of the synthetic data we created, we performed clustering based on two criteria: (1) topics that trigger negative emotions in adolescents and (2) negative emotions and symptoms outlined in the DSM-5\footnote{\url{https://www.mdcalc.com/calc/10195/dsm-5-criteria-major-depressive-disorder}}, a widely used framework for assessing and diagnosing mental disorders \cite{lee2023methodology}.

\subsection{Topic-Based Classification of Adolescent Negative Thinking}

The Korea National Youth Policy Institute (NYPI\footnote{\url{https://www.nypi.re.kr/}}), under the Ministry of Gender Equality and Family, categorized adolescents' concerns into five areas: (1) academic and career concerns, (2) relationships (friendships, romance, bullying), (3) physical and mental health, (4) family issues, and (5) appearance and self-image.

To assess the alignment of our synthetic data on adolescent negative thinking with these categories, we applied K-means clustering, an unsupervised machine learning algorithm that partitions data into distinct groups, to keywords extracted from 37,124 adolescents' questions (Section~3.1). Specifically, K-means clustering was used to define hierarchical relationships and classifications, grouping the data into the five predefined subject clusters, each of which was assigned sub-keywords based on relevance. As a result, a dictionary with five topics and 120 keywords was created, as shown in Table~\ref{tab:topic-keywords} in Appendix~\ref{appendixF}.

We identified the most frequent keywords for each topic. The top topic was academic performance and career concerns, with 99,076 instances (36.9\%), followed by relationships (73,586, 27.4\%), physical and mental health (71,249, 26.5\%), family issues (20,532, 7.6\%), and appearance and self-image (4,007, 1.5\%).

\subsection{DSM-5 Based Classification of Adolescent Negative Thinking}

Cognitive distortions can contribute to depression, so we examined the nine categories of the DSM-5 to determine whether a significant relationship exists. To explore this, we analyzed 37,124 adolescents' questions (Section~3.1) and identified DSM-5-related word distributions using NLTK text mining\footnote{\url{https://www.nltk.org/book/ch07.html}} (Section~3.9.1). We used NLTK to build a keyword-topic dictionary based on DSM-5 criteria, extracting synonyms for DSM-5 symptoms with WordNet, identifying related keywords, and using LDA topic modeling to derive and compare topics with the DSM-5 criteria. These distributions were then used to create dictionaries for DSM classification, resulting in nine categories and 121 keywords, as shown in Table~\ref{tab:dsm-keywords} in Appendix~\ref{appendixF}.

For keyword mapping, we used our dataset of 108,717 synthesized data points (Section~3.5.2), allowing multiple keywords per data point. For DSM-based keyword mapping, 69,290 data points (63.7\%) were successfully mapped, with 115 unique keywords assigned 1,335,337 times. For negative emotion-triggering topic-based mapping, 103,183 data points (94.9\%) were successfully mapped, with 129 unique keywords assigned 268,450 times.

Among the DSM-5 symptom categories, five out of nine categories appeared more than 15,000 times. The most frequent keyword was B. Loss of interest or pleasure (321,157 occurrences, 23.8\%), followed by H. Decreased concentration (25,580 occurrences, 18.9\%), A. Depressed mood (25,258 occurrences, 18.7\%), D. Insomnia or hypersomnia (24,864 occurrences, 18.4\%), and E. Psychomotor agitation or retardation (15,235 occurrences, 11.3\%).

\begin{figure}[ht]
  \centering
  \includegraphics[width=0.95\linewidth]{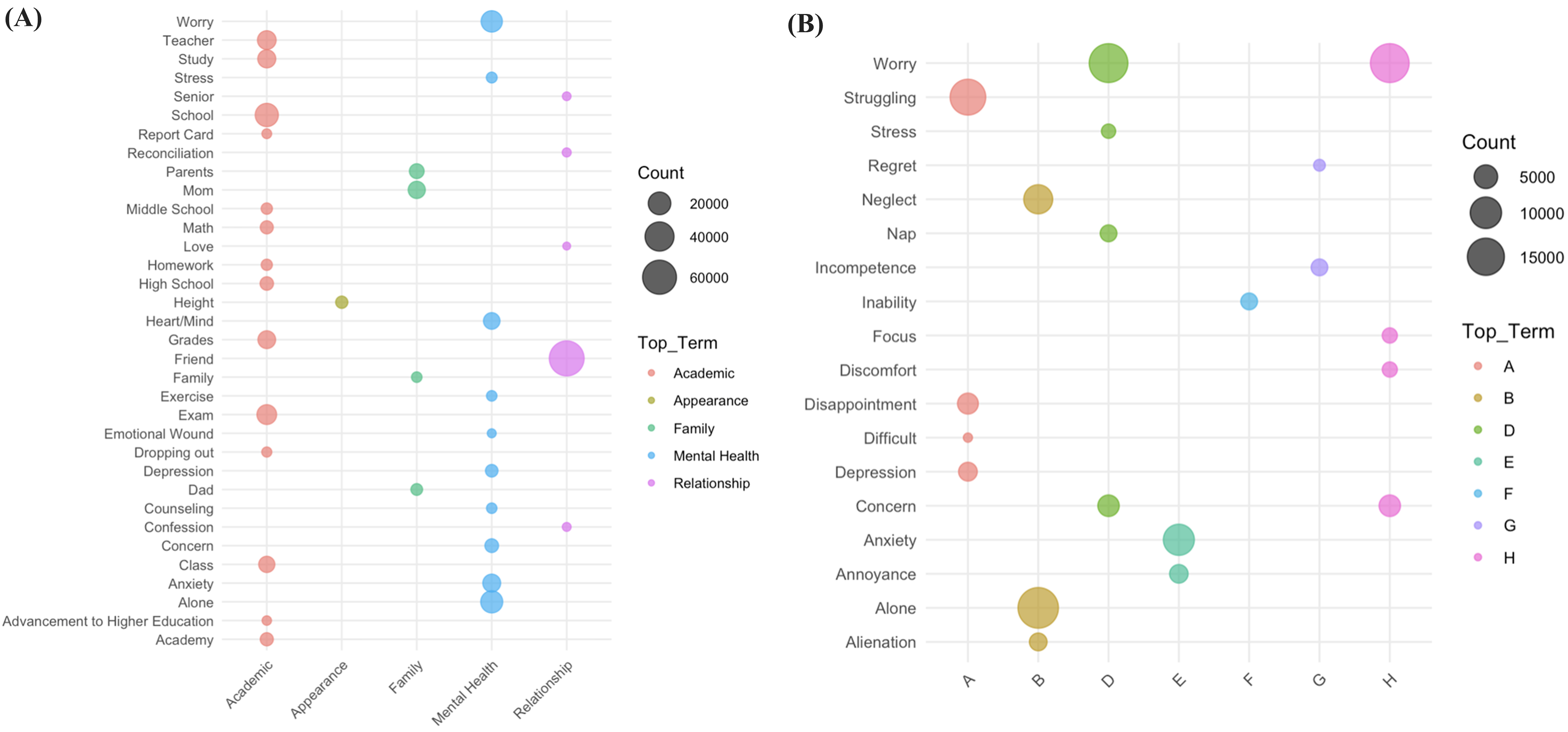}
  \caption{Cluster distribution of high-frequency keywords: (A) Negative emotion-triggering topics and (B) DSM-5 symptom keywords ($\geq$1,000 occurrences).}
  \label{fig:clustering_keywords}
\end{figure}

We found 34 keywords (Table~\ref{tab:topic-mapping} in Appendix~\ref{appendixF} for cognitive distortion-triggering topics and 20 (Table~\ref{tab:dsm-mapping} in Appendix~\ref{appendixF}) for DSM-5 categories, each with a frequency of 1,000 or more, as listed in Figure~\ref{fig:clustering_keywords}.

The generated synthetic data mainly highlighted academic and career stress, along with social conflicts like friendships and romantic relationships, while underrepresenting appearance and self-image issues. Additionally, its cognitive distortions were closely linked to five of the nine DSM-5 depression symptom keywords.

\section{Evaluation}

We evaluate the quality of two types of synthetic data: data generated from clearly identified cognitive distortions (Section~3.5.1) and data generated to address cognitive distortion imbalances (Section~3.5.2). The evaluation was conducted independently using three evaluation criteria, with both LLMs and human assessments.

\subsection{Evaluation Criteria}

We evaluated the generated synthetic data using three criteria: (1) Consistency, (2) Accuracy, and (3) Fluency. Scores ranged from 1 to 3, with 1 indicating ``inappropriate'' and 3 indicating ``highly appropriate''. Consistency checked if the cognitive distortion was logically maintained between the original and synthetic data. Accuracy assessed whether the labeled cognitive distortion matched the correct classification. Fluency evaluated how natural, grammatically correct, and easy to read the sentences were. The prompts used for these criteria are provided in Appendix~\ref{appendixH}.

\subsection{Comparison of LLMs and Human Evaluations Across Criteria}

To ensure objectivity, the model generating synthetic data was excluded from evaluation. Two other models independently scored the data, averaging their scores for the final result. Evaluation parameters for the three LLMs are in Table~\ref{tab:llm-hyperparams}, Appendix~\ref{appendixE}.

For human evaluation, 50 or 100 synthetic samples per distortion were randomly selected, totaling 900 samples. Two psychology experts independently assessed them using the same criteria as LLM evaluation, with a Cohen's kappa of 0.78 indicating substantial agreement.

Table~\ref{tab:evaluation_results} summarizes the evaluation results from both LLMs and humans, highlighting the differences between the two types of synthetic data (Section~3.5)—cognitive clarification and cognitive balancing—across the three criteria. Detailed results for each model are in Table~\ref{tab:llm-assessment-details}, Appendix~\ref{appendixD}.

Human evaluation scores were lower across all criteria except fluency, with accuracy showing the largest gap. This difference stems from LLMs' strength in detecting explicit text patterns while struggling with the implicit reasoning essential for cognitive distortion evaluation, highlighting their limitations. Table~\ref{tab:expert_feedback} provides detailed expert feedback.

Regarding the two synthetic data generation methods, in the LLM evaluation, the cognitive clarification method scored 0.1 to 0.3 points higher on all criteria than the cognitive balancing method. However, in the human evaluation, only the cognitive balancing method showed higher accuracy.

\begin{table}[htbp]
\centering
\renewcommand{\arraystretch}{1.2}
\setlength{\tabcolsep}{8pt}
\begin{tabular}{lcccc}
\toprule
\textbf{Criteria} & \multicolumn{2}{c}{\textbf{LLMs Evaluation}} & \multicolumn{2}{c}{\textbf{Human Evaluation}} \\
\cmidrule(r){2-3} \cmidrule(r){4-5}
& Cognitive Clarification & Cognitive Balancing & Cognitive Clarification & Cognitive Balancing \\
\midrule
Consistency & 2.400 $\pm$ 0.232 & 2.105 $\pm$ 0.173 & 2.254 & 2.160 \\
Accuracy    & 2.708 $\pm$ 0.177 & 2.416 $\pm$ 0.270 & 2.322 & 2.738 \\
Fluency     & 2.655 $\pm$ 0.219 & 2.529 $\pm$ 0.223 & 2.904 & 2.690 \\
\bottomrule
\end{tabular}

\setlength{\abovecaptionskip}{10pt} 
\caption{Evaluation results of synthetic data by LLMs and humans. 
The LLMs evaluation (left) reports mean ± standard deviation scores assigned by three models, 
where the standard deviation represents variations across models. 
The human evaluation (right) presents the average scores given by two experts after cross-validation.}
\label{tab:evaluation_results}
\end{table}

\subsection{Comparison of LLM and Human Performance in Cognitive Distortion Classification}

To further analyze the differences between LLM-based and human evaluations, we compared the scores for each cognitive distortion. Table~\ref{tab:comparison_llm_human} presents comparative results, highlighting key discrepancies between the two evaluation methods. Scores were compared between LLM and human evaluations, with the higher values in bold. The 'Difference' column shows score gaps, with differences of 0.4 or greater also bolded.

The average LLMs evaluation scores were 2.287 for consistency, 2.582 for accuracy, and 2.598 for fluency, while the average human evaluation scores were 2.278 for consistency, 2.558 for accuracy, and 2.815 for fluency. Fluency was higher in human evaluation, whereas consistency and accuracy showed no significant difference, though human scores were slightly lower overall. The higher fluency score in human evaluation is likely because LLMs assessed synthetically generated sentences, which were naturally structured and free of pauses.

In the evaluation of cognitive distortions by type, human scores were lower than those of LLMs in some cases, particularly in accuracy. For example, scores for "Emotional Reasoning" (2.887 vs. 2.200) and "Magnification and Minimization" (2.531 vs. 2.100) showed notable differences. This discrepancy may be because LLMs excel at detecting clear linguistic patterns, such as "Should Statements," "Labeling," and "Discounting the Positive." However, human evaluation tends to be more reliable for distortions requiring inferential reasoning, such as "Mental Filtering" and "Magnification and Minimization," since these rely on deeper contextual understanding.

These findings highlight that LLMs rely more on explicit linguistic patterns, whereas human evaluators consider deeper contextual reasoning, which may impact their ability to identify distortions that require implicit inference.

\begin{table}[htbp]
\centering
\renewcommand{\arraystretch}{1.3}
\setlength{\tabcolsep}{6pt}
\begin{tabular}{m{4.0cm} m{9.5cm}}
\toprule
\textbf{Original Distress Question} & 
\textit{"Whenever I meet my cousins, my mom asks me why I’m not tall. Even my friends, who are shorter than me, stay up past 2 a.m., and when I see them grow taller, I can only think about why I’m not tall."} \\
\midrule
\textbf{[Type of Cognitive Distortion] Synthetic Data} & 
\textbf{["Should" Statement]} \newline
\textit{"My mom often compares me to her cousins and says, 'Why are you so short?' I don’t understand why I’m the only one who’s so short when all my friends are growing taller. Both my mom and dad are tall, but I feel like something is wrong with being short."} \\
\midrule
\textbf{Commands from Expert} & 
The expert provided two points in the accuracy evaluation of the cognitive distortion type, and chose mental filtering rather than should statement. 

\textit{The belief that one should be tall (should statement) usually comes from parents. In this article, we confirmed that the fact that the mother is not tall triggers anxiety. However, this presupposes that the individual has negative thoughts (mental filtering), as she believes she will not grow taller. While the should statement seems to be the main issue in literal terms, mental filtering—an error in self-judgment—is considered the primary cognitive distortion.} \\
\bottomrule
\end{tabular}

\setlength{\abovecaptionskip}{10pt} 
\caption{Expert analysis of a case with synthetic data accuracy score of 2: Explanations for LLM misclassification.}
\label{tab:expert_feedback}
\end{table}

\begin{table}[htbp]
\centering
\renewcommand{\arraystretch}{1.2}
\setlength{\tabcolsep}{6pt}
\begin{tabular}{lccccccccc}
\toprule
\textbf{Cognitive Distortion Type} & \multicolumn{3}{c}{\textbf{LLMs Evaluation}} & \multicolumn{3}{c}{\textbf{Human Evaluation}} & \multicolumn{3}{c}{\textbf{Difference}} \\
\cmidrule(r){2-4} \cmidrule(r){5-7} \cmidrule(r){8-10}
& Cos & Acc & Flu & Cos & Acc & Flu & Cos & Acc & Flu \\
\midrule
All-or-Nothing Thinking* & 2.203 & \textbf{2.607} & 2.470 & \textbf{2.610} & 2.590 & \textbf{2.730} & \textbf{0.407} & 0.017 & 0.260 \\
Overgeneralization* & \textbf{2.287} & \textbf{2.767} & 2.609 & 2.280 & 2.520 & \textbf{2.860} & 0.007 & 0.247 & 0.251 \\
Mental Filtering* & \textbf{2.247} & \textbf{2.677} & 2.578 & 2.480 & 2.460 & \textbf{2.830} & 0.233 & 0.217 & 0.252 \\
Discounting the Positive & \textbf{2.153} & 2.240 & 2.640 & 2.120 & \textbf{2.710} & \textbf{2.880} & 0.033 & 0.470 & 0.240 \\
Jumping to Conclusions & 2.279 & 2.361 & 2.550 & \textbf{2.560} & \textbf{2.890} & \textbf{2.840} & 0.281 & 0.529 & 0.290 \\
Magnification and Minimization* & 2.212 & \textbf{2.531} & 2.625 & \textbf{2.330} & 2.100 & \textbf{2.730} & 0.118 & 0.431 & 0.105 \\
Emotional Reasoning* & \textbf{2.624} & \textbf{2.887} & 2.713 & 2.020 & 2.200 & \textbf{2.880} & \textbf{0.604} & \textbf{0.687} & 0.167 \\
Should Statements & \textbf{2.315} & 2.562 & 2.654 & 2.110 & \textbf{2.600} & \textbf{2.770} & 0.205 & 0.038 & 0.116 \\
Labeling & \textbf{2.309} & \textbf{2.563} & 2.499 & 2.380 & 2.700 & \textbf{2.770} & 0.071 & 0.137 & 0.271 \\
Personalization & \textbf{2.250} & 2.632 & 2.648 & 1.890 & \textbf{2.810} & \textbf{2.860} & 0.360 & 0.178 & 0.212 \\
\midrule
Total mean & 2.287 & 2.582 & 2.598 & 2.278 & 2.558 & 2.815 & 0.231 & 0.295 & 0.216 \\
\bottomrule
\end{tabular}

\setlength{\abovecaptionskip}{10pt} 
\caption{Comparative evaluation of cognitive distortions by LLMs and humans: Cos (Consistency), Acc (Accuracy), and Flu (Fluency). *Types of cognitive distortions easily detected by LLMs.}
\label{tab:comparison_llm_human}
\end{table}

\section{Conclusion and Future Work}

We developed KoACD, a dataset of cognitive distortions in Korean adolescents, overcoming the limitations of small-scale datasets focused on English-speaking adults. KoACD offers a balanced representation of cognitive distortions through the creation of synthetic data. To our knowledge, it is the first dataset specifically designed for Korean adolescents.

We introduced a multi-LLM negotiation method to improve the objectivity and accuracy of the synthetic data. By using multiple LLMs to negotiate and refine cognitive distortion labels, we minimized biases and enhanced data quality. Expert and LLM evaluations confirmed that LLMs performed well when clear linguistic cues were present, while human evaluators showed higher accuracy in context-dependent situations. Discrepancies between LLMs and human evaluations highlighted the LLMs' reliance on superficial linguistic patterns.

Future work will focus on fine-tuning models with adolescent-specific data to enhance contextual understanding of cognitive distortions. Additionally, we aim to improve LLM performance by developing algorithms that better distinguish cognitive distortions, mitigating biases toward specific types and enhancing both balance and accuracy in detection.

\section{Limitations}

We recognize that there are some limitations to the methods for detecting cognitive distortions and to the KoACD dataset:

\paragraph{Cognitive Distortion Classification} 
We assigned the most appropriate cognitive distortion to each question, but some questions may involve multiple distortions simultaneously. The boundaries between some types of distortions are blurred, making classification challenging and leading to potential discrepancies between the model and human raters. To address these issues, a multi-label classification method and more refined criteria are needed.

\paragraph{Multi-LLMs Negotiation Methods} 
We designed the LLMs to alternate between Analyzer and Evaluator roles, but the results can vary depending on the model used. Therefore, negotiation results with different LLMs should also be considered. Additionally, discrepancies between analysts and evaluators sometimes result in data being classified as "Unknown," even after five rounds of negotiation, due to the inability to fit the data within the ten cognitive distortion categories. Interpretation of such data is essential, and further research is needed to develop more accurate detection methodologies.

\paragraph{LLMs and Human Evaluation} 
While the KoACD is a large dataset, the amount of data reviewed by human raters is relatively small. Although human raters excel at considering context for accurate judgments, subjectivity in the evaluation process and inconsistency due to differing standards among raters may arise. Future research should focus on securing more human evaluation data and developing more precise evaluation standards to increase reliability.

\section{Ethical Considerations}

In this study, we collected publicly accessible data from NAVER Knowledge iN, and users participate anonymously on the platform. We only used publicly available data in the course of our research and did not interact directly with NAVER Knowledge iN users.

We have identified that the data collection process may include various inappropriate topics, such as hate speech, violence, sexual content, and profanity. Accordingly, we have attempted to exclude such data as much as possible by applying strict filtering criteria, including keyword-based filtering and manual inspection of sampled posts. However, we cannot completely rule out the possibility that some inappropriate content may be included in the data.

We are aware of the risk that AI models may be trained on inappropriate data and produce biased or unethical results. Therefore, it is important to continuously monitor the ethical use of AI models and improve filtering techniques to address this risk.

All processed and derived datasets are intended solely for research purposes and must not be used for commercial or any other non-research purposes. Their use is consistent with the original access conditions of NAVER Knowledge iN.

To ensure the reliability of the cognitive distortion annotations, two psychological experts—each holding a master's degree in clinical or counseling psychology and possessing over five years of relevant experience—independently reviewed and validated 900 samples from the dataset. The experts assessed the appropriateness of the assigned cognitive distortion types based on predefined guidelines established for this study.

All data samples were fully anonymized prior to evaluation to eliminate any risk of personal identification. No personal information was collected from either the dataset or the expert reviewers.

The experts voluntarily participated in the annotation task after being fully informed of the study's purpose and procedures. They were compensated with 300,000 KRW, an amount determined to reflect fair payment based on standard professional rates and estimated task duration.

The annotation process was conducted under strict confidentiality, and all procedures adhered to ethical standards for research involving domain experts.

All experts provided informed consent to participate in this study, and we explained how the annotation evaluation would be conducted and how their feedback would be used for research purposes.

\bibliographystyle{unsrt}  
\bibliography{references}  

\appendix

\section{Distribution of Data Sources}
\label{appendix:data-distribution}
We collected 69,925 questions and answers from five major organizations and services specializing in adolescent counseling on NAVER Knowledge iN. Table~\ref{tab:data-distribution} shows the data collection status by organization and year, along with the distribution of questions collected from 2011 to 2024.

\begin{table}[htbp]
\centering
\renewcommand{\arraystretch}{1.2}
\setlength{\tabcolsep}{6pt}
\begin{tabular}{lrrrr}
\hline
\textbf{Institution} & \textbf{2011--2015} & \textbf{2016--2020} & \textbf{2021--2024} & \textbf{Total} \\
\hline
\begin{tabular}[c]{@{}l@{}}여성가족부·한국청소년상담복지개발원 \\ 
Korea Youth Counseling \& Welfare Institute\end{tabular} & 17,479 & 19,465 & 15,699 & 52,643 \\
\begin{tabular}[c]{@{}l@{}}인천시 청소년 지원센터 \\ 
Incheon Youth Support Center\end{tabular} & 1,357 & 1,355 & -- & 2,712 \\
\begin{tabular}[c]{@{}l@{}}울산시 청소년 지원센터 \\ 
Ulsan Youth Support Center\end{tabular} & 1,655 & 2,360 & 1,518 & 5,533 \\
\begin{tabular}[c]{@{}l@{}}경기도 청소년 지원센터 \\ 
Gyeonggi Youth Support Center\end{tabular} & 6,862 & -- & 53 & 6,915 \\
\begin{tabular}[c]{@{}l@{}}청소년 모바일 상담센터 \\ 
Youth Mobile Counseling Center\end{tabular} & -- & 407 & 1,715 & 2,122 \\
\hline
\textbf{Overall Total} & \textbf{27,353} & \textbf{23,587} & \textbf{18,985} & \textbf{69,925} \\
\hline
\end{tabular}

\setlength{\abovecaptionskip}{10pt} 
\caption{Distribution of Q\&A and worry Q\&A data by institution and year}
\label{tab:data-distribution}
\end{table}

\section{Data Preprocessing Details}
\label{appendixB}
The collected data was refined and pre-processed to ensure relevance. The following criteria were applied to remove misaligned data:

\begin{enumerate}
    \item Non-adolescent questions: To exclude questions written by elementary school students or adults, we applied keyword-based filtering, resulting in the removal of 14,075 questions.

    \item Inappropriate content: A total of 7,397 questions containing inappropriate sexual content were removed to maintain alignment with the research scope.

    \item Lack of specificity: To eliminate vague questions that hinder meaningful analysis, 9,240 questions with 15 words or fewer in the detailed worry column were deleted.

    \item Duplicate entries: To ensure data uniqueness and prevent redundancy, 2,089 duplicate questions were removed.
\end{enumerate}

After applying the above criteria for pre-processing, 37,124 data points were selected and used in the study.

\section{Changes in Cognitive Distortion Classification Through Negotiation}
\label{appendixC}
We analyze the distribution of data based on the number of negotiation rounds required to determine cognitive distortions. Table~\ref{tab:negotiation-rounds} presents the count of instances finalized at each round, illustrating how much data was classified early versus how much required additional rounds. The cumulative percentage represents the proportion of data for which cognitive distortion classification was completed at each round.

\begin{table}[htbp]
\centering
\renewcommand{\arraystretch}{1.2}
\setlength{\tabcolsep}{6pt}
\begin{tabular}{lrrr}
\toprule
\textbf{Round} & \textbf{Turn 1} & \textbf{Turn 2} & \textbf{Total (Cumulative \%)} \\
\midrule
Round 1 & 17,694 & 1,841 & 17,694 (51\%) \\
Round 2 & 6,132  & 361   & 6,493 (75\%)  \\
Round 3 & 4,240  & 57    & 4,297 (87\%)  \\
Round 4 & 1,243  & 37    & 1,280 (91\%)  \\
Round 5 & 784    & 4,735 & 5,519 (100\%) \\
\bottomrule
\end{tabular}

\setlength{\abovecaptionskip}{10pt}  
\caption{Turn counts across negotiation rounds}
\label{tab:negotiation-rounds}
\end{table}

\section{Detailed Evaluation Results of LLM-Based Assessment}
\label{appendixD}

This appendix presents the detailed evaluation results of the LLM-based assessment for the two synthetic data generation methods: Cognitive Clarification and Cognitive Balancing. Each model's performance was assessed based on three criteria—Consistency (Cos), Accuracy (Acc), and Fluency (Flu)—using independent evaluations by Gemini 1.5 Flash, GPT-4o mini, and Claude 3 Haiku, as shown in Table~\ref{tab:llm-assessment-details}.

\begin{table}[htbp]
\centering
\renewcommand{\arraystretch}{1.2}
\setlength{\tabcolsep}{6pt}
\resizebox{\textwidth}{!}{%
\begin{tabular}{llccccccccc}
\toprule
\multirow{2}{*}{\textbf{Generation Method}} & \multirow{2}{*}{\textbf{Generation Model}} 
& \multicolumn{3}{c}{\textbf{Gemini-1.5-flash}} 
& \multicolumn{3}{c}{\textbf{GPT-4o mini}} 
& \multicolumn{3}{c}{\textbf{Claude-3-haiku}} \\
\cmidrule(lr){3-5} \cmidrule(lr){6-8} \cmidrule(lr){9-11}
& & Cos & Acc & Flu & Cos & Acc & Flu & Cos & Acc & Flu \\
\midrule
\multirow{3}{*}{Cognitive Clarification} 
& Gemini-1.5-flash & - & - & - & 2.596 & 2.929 & 2.948 & 2.400 & 2.779 & 2.416 \\
& GPT-4o mini      & 2.142 & 2.498 & 2.638 & - & - & - & 2.508 & 2.774 & 2.472 \\
& Claude-3-haiku   & 2.150 & 2.519 & 2.643 & 2.606 & 2.754 & 2.814 & - & - & - \\
\midrule
\multirow{3}{*}{Cognitive Balancing} 
& Gemini-1.5-flash & - & - & - & 2.253 & 2.718 & 2.740 & 2.134 & 2.589 & 2.298 \\
& GPT-4o mini      & 1.882 & 2.090 & 2.515 & - & - & - & 2.111 & 2.333 & 2.312 \\
& Claude-3-haiku   & 1.966 & 2.162 & 2.547 & 2.284 & 2.604 & 2.760 & - & - & - \\
\bottomrule
\end{tabular}
}

\setlength{\abovecaptionskip}{10pt}  
\caption{Detailed evaluation results of synthetic data: Cos (Consistency), Acc (Accuracy), and Flu (Fluency)}
\label{tab:llm-assessment-details}
\end{table}

\section{Hyperparameters of LLMs Models}
\label{appendixE}

We utilized Claude-3 Haiku, Gemini-1.5 Flash, and GPT-4o Mini at different stages of this study, summarizing the hyperparameters used at each step. Table~\ref{tab:llm-hyperparams}(A) presents the hyperparameters for the negotiation process and independent evaluation, with Gemini-1.5 Flash and GPT-4o Mini used during the negotiation, and Claude-3 Haiku employed for independent evaluation. Table~\ref{tab:llm-hyperparams}(B) summarizes the hyperparameters for synthetic data generation, while Table~\ref{tab:llm-hyperparams}(C) outlines the hyperparameters used for evaluating synthetic data. The hyperparameters were empirically determined through multiple experiments, focusing on maximizing output diversity, ensuring reproducibility, and guaranteeing sufficient output size.

\begin{table}[htbp]
\centering
\renewcommand{\arraystretch}{1.2}
\setlength{\tabcolsep}{8pt}
\begin{tabular}{llccc}
\toprule
\textbf{Methodology} & \textbf{Model} & \textbf{Temperature} & \textbf{Max Tokens} & \textbf{Top-p} \\
\midrule
\multirow{3}{*}{(A) Negotiation process} 
& Gemini-1.5 Flash & 0.5 & 1,024 & 0.9 \\
& GPT-4o mini       & 0.5 & 1,024 & 0.9 \\
& Claude-3 Haiku    & 0.5 & 1,024 & 0.9 \\
\midrule
\multirow{3}{*}{(B) Synthetic data generation} 
& Gemini-1.5 Flash & 1.0 & 1,024 & 0.9 \\
& GPT-4o mini       & 1.0 & 1,024 & 0.9 \\
& Claude-3 Haiku    & 1.0 & 1,024 & 0.9 \\
\midrule
\multirow{3}{*}{(C) Evaluation} 
& Gemini-1.5 Flash & 0.5 & 512 & 0.9 \\
& GPT-4o mini       & 0.5 & 512 & 0.9 \\
& Claude-3 Haiku    & 0.5 & 512 & 0.9 \\
\bottomrule
\end{tabular}

\setlength{\abovecaptionskip}{10pt}  
\caption{Hyperparameters for negotiation process, synthetic data generation, and evaluation}
\label{tab:llm-hyperparams}
\end{table}

\clearpage
\section{Validating Synthetic Data with Clustering}
\label{appendixF}

To validate the synthetic data, we conducted clustering based on two criteria: 
(1) topics that elicit cognitive distortion in adolescents and 
(2) negative emotions and symptoms from the DSM-5.

To perform clustering, we first created mapping dictionaries for each criterion. 
Table~\ref{tab:topic-keywords} lists keywords for cognitive distortion topics in adolescents, 
and Table~\ref{tab:dsm-keywords} shows DSM-5 depression symptom categories with related keywords.

We discovered keywords mapped to each category based on the topic-based mapping (Table~\ref{tab:topic-mapping}) 
and DSM-5 symptom-based mapping (Table~\ref{tab:dsm-mapping}) of the synthetic data in the mapping dictionary. 
Only keywords with a mapping frequency of over 1,000 were selected, and the results were checked with the keyword 
in English, Korean, and the mapping frequency.

\begin{table}[htbp]
\renewcommand{\arraystretch}{1.2}
\setlength{\tabcolsep}{8pt}

\begin{tabular}{p{5cm}p{10cm}}
\toprule
\textbf{Topic (Korean, n)} & 
\multicolumn{1}{c}{\textbf{List of Keywords (Korean)}} \\
\midrule
Academic performance and career concerns \newline (학업 성취도 및 진로, n=31) & 
Academics (학업), Academy (학원), Advancement to Higher Education (진학), Class (수업), Club (동아리), College Entrance Exam (입시), College Entrance Exam (수능), Dropping out (자퇴), English (영어), English Academy (영어학원), Enrollment (재학), Exam (시험), Final Exam (기말고사), GED (검정고시), Grades (성적), Harass/Bully (괴롭히다), High School (고등학교), Homework (숙제), Math (수학), Middle School (중학교), Midterm Exam (중간고사), Mock Exam (모의고사), Private Tutoring (과외), Rank (등급), Report Card (성적표), Retaking the College Entrance Exam (재수), Scholarship (장학금), School (학교), School Life (학교생활), School Record (내신), Specialized High School (특성화고), Study (공부), Teacher (선생님) \\
\midrule
Friendships, romantic relationships, and interpersonal relationships \newline (우정, 연애, 대인관계, n=19) &
Acquaintance (지인), Best Friend (단짝), Boyfriend (남자친구), Boyfriend (남친), Break up (헤어지다), Bullying (왕따), Close Friend (친한친구), Confession (고백), Crush (짝사랑), Dating (사귀다), Friend (친구), Friendship (친구사이), Girlfriend (여자친구), Heartbreak (실연, 마음의 상처), Jealousy (질투), Love (사랑), Loyalty (우정, 의리) \\
\midrule
Physical and mental health \newline (신체적, 정신적 건강, n=39) &
Alone (혼자), Anxiety (불안), Binge Eating Disorder (폭식증), Comfort (위로), Confidence (자신감), Concern (고민), Counseling (상담), Counselor (상담사), Depression (우울), Depression (우울증), Domestic Violence (가정폭력), Emotional Wound (상처), Exercise (운동), Fatigue (피로), Guilt (죄책감), Headache (두통), Heart/Mind (마음), Inferiority Complex (열등감), Inner Self (내면), Insomnia (불면증), Loneliness (외로움), Mental Illness (정신병), Mental Strength (멘탈), Obesity (비만), Psychotherapy (심리치료), Running Away (가출), School Bullying (학폭), School Violence (학교폭력), Self-esteem (자존감), Sleep Disorder (수면 장애), Stress (스트레스), Therapy (치료), Unconscious Mind (무의식), Violence (폭력), Worry (걱정) \\
\midrule
Family issues \newline (가족 문제, n=20) &
Dad (아빠), Divorce (이혼), Domestic Conflict (가정 불화), Domestic Violence (가정폭력), Estrangement (소원함), Family (가족), Family Conflict (가족 갈등), Family History (가족사), Father (아버지), Financial Issues (경제적 문제), Home/Family Environment (가정), Mom (엄마), Mother (어머니), Neglect (방임), Older Sister (누나), Older Sister (언니), Parents (부모님), Younger Brother (남동생), Younger Sibling (동생), Younger Sister (여동생) \\
\midrule
Appearance and self-image \newline (외모 및 이미지, n=11) &
Acne (여드름), Appearance (외모), Body Shape (몸매), Bulking Up (벌크업), Diet (다이어트), Facial Features (얼굴 생김새), Height (키), Makeup (메이크업), Muscle (근육), Plastic Surgery (성형), Skin (피부) \\
\bottomrule
\end{tabular}

\setlength{\abovecaptionskip}{10pt}  
\caption{List of Keywords of Negative Emotion-Triggering Topics in Adolescents}
\label{tab:topic-keywords}
\end{table}

\begin{table}[H]
\centering
\renewcommand{\arraystretch}{1.3} 
\setlength{\tabcolsep}{6pt}      

\footnotesize
\begin{tabular}{m{6cm} m{9cm}}
\toprule
\textbf{DSM-5 Depression Symptom Class \newline (Korean, n)} &
\multicolumn{1}{c}{\textbf{List of Symptom or Emotion (Korean)}} \\
\midrule

A. Depressed mood \newline (우울한 기분, n=21) &
Crying (울다), Depression (우울, 우울증), Despair (절망), Disappointment (실망), Emptiness (공허, 허탈), Frustration (좌절), Guilt (죄책감), Hard (힘들다), Heartache (상심), Pain (고통), Sad (슬프다), Scared (무서운, 겁나는), Suffering (괴로움), Tough (힘들), Unhappiness (불행), Upset (화나다, 속상함), Worthlessness (무가치함) \\
\midrule
B. Loss of interest/pleasure \newline (흥미 또는 즐거움의 상실, n=14) &
Alienated (소외), Alone (혼자, 홀로), Apathy (냉담), Bore (지루함), Bullying (따돌림), Unpleasant (불쾌함), Ignored (무시), Indifference (무관심), Isolated (고립), Loneliness (외로움), Lonely (외로워), Meaningless (무의미함), Disinterest (흥미 없음) \\
\midrule
C. Weight loss or gain \newline (체중 감소 또는 증가, n=9) &
Appetite (식욕), Binge Eating (폭식), Body (몸매), Diet (다이어트, 식단), Fat (살찌다), Loss of Appetite (식욕 감퇴), Nausea (매스꺼움), Weight (체중) \\
\midrule
D. Insomnia or hypersomnia \newline (불면증 또는 과다수면, n=15) &
Daytime Fatigue (주간 피로), Hypersomnia (과다수면), Insomnia (불면, 불면증), Restless Sleep (뒤척임), Sleep (수면, 잠), Sleep Deprivation (수면 부족), Sleep Disorder (수면장애), Sleep Patterns (수면 패턴), Sleepiness (졸음), Sleeping Pills (수면제), Stress (스트레스), Worry (고민, 걱정) \\
\midrule
E. Psychomotor agitation or retardation \newline (정신운동 초조 또는 지연, n=13) &
Anger (분노), Anxiety (불안, 불안감), Anxiety disorder (불안 장애), Irritability (과민, 과민성, 짜증), Nervousness (초조, 신경질), Obsessive (강박증), Obsessive-compulsive disorder (강박장애), Sensitive (예민, 예민한), Tension (긴장, 긴장감) \\
\midrule
F. Fatigue \newline (피로감, n=17) &
Dejected (낙담, 허탈), Empty (공허), Exhausted (지치다, 탈진), Fatigued (피로), Helpless (무력감), Incompetence (무능, 능력 부족), Inferiority (열등감, 자신감 부족), Lethargy (무기력, 무기력증), Powerless (무기력한, 힘이 없는), Tired (피곤, 피곤함) \\
\midrule
G. Inappropriate guilt \newline (부적절한 죄책감, n=11) &
Guilt (죄책감), Helplessness (무력감), Incompetence (무능, 무능함, 무능력), Inferiority (열등감), Regret (후회), Self-blame (자책), Shame (수치, 창피, 수치심) \\
\midrule
H. Decreased concentration \newline (집중력 저하, n=12) &
Concentration (집중, 집중력), Concern (염려, 우려, 고민), Confusion (혼란), Distracted (산만함, 주의 산만), Discomfort/Inconvenience (불편, 불편함), Forgetfulness (건망증), Judgment (판단) \\
\midrule
I. Thoughts of suicide \newline (자살 사고, n=9) &
Death (죽음), Desperation (절박함, 절망), Die (죽다), Fear (두려움), Self-harm (자해), Suicide (자살), Suicidal Ideation (자살 충동, 자살 사고) \\
\bottomrule
\end{tabular}

\setlength{\abovecaptionskip}{10pt}  
\caption{DSM-5 Depression Symptom related Classes and Keywords}
\label{tab:dsm-keywords}
\normalsize
\end{table}

\begin{table}[H]
\centering
\renewcommand{\arraystretch}{1.2}
\setlength{\tabcolsep}{6pt}
\begin{tabular}{llll}
\toprule
\textbf{Topic} & \textbf{Keyword\_KR} & \textbf{Keyword\_ENG} & \textbf{Count} \\
\midrule
Relationship & 친구 & Friend & 64,041 \\
Academic & 학교 & School & 21,298 \\
Mental Health & 혼자 & Alone & 18,482 \\
Mental Health & 걱정 & Worry & 16,636 \\
Academic & 시험 & Exam & 13,168 \\
Academic & 선생님 & Teacher & 11,163 \\
Academic & 공부 & Study & 9,987 \\
Mental Health & 불안 & Anxiety & 9,716 \\
Academic & 성적 & Grades & 9,407 \\
Family & 엄마 & Mom & 8,458 \\
Mental Health & 마음 & Heart/Mind & 7,543 \\
Academic & 수업 & Class & 6,987 \\
Family & 부모님 & Parents & 5,121 \\
Mental Health & 고민 & Concern & 3,966 \\
Academic & 고등학교 & High School & 3,725 \\
Academic & 학원 & Academy & 3,409 \\
Academic & 수학 & Math & 3,347 \\
Mental Health & 우울 & Depression & 2,927 \\
Appearance & 키 & Height & 2,690 \\
Family & 아빠 & Dad & 2,312 \\
Academic & 중학교 & Middle School & 2,065 \\
Academic & 숙제 & Homework & 1,992 \\
Mental Health & 스트레스 & Stress & 1,798 \\
Mental Health & 운동 & Exercise & 1,639 \\
Family & 가족 & Family & 1,636 \\
Mental Health & 상담 & Counseling & 1,619 \\
Academic & 자퇴 & Dropping out & 1,458 \\
Academic & 성적표 & Report Card & 1,343 \\
Academic & 진학 & Advancement to Higher Education & 1,241 \\
Relationship & 화해 & Reconciliation & 1,148 \\
Mental Health & 상처 & Emotional Wound & 1,125 \\
Relationship & 고백 & Confession & 1,113 \\
Relationship & 선배 & Senior & 1,086 \\
Relationship & 사랑 & Love & 1,025 \\
\bottomrule
\end{tabular}
\setlength{\abovecaptionskip}{12pt}  
\setlength{\belowcaptionskip}{0pt}   
\caption{Frequency Distribution of 34 Keywords Across Topics}
\label{tab:topic-mapping}
\end{table}

\begin{table}[H]
\centering
\renewcommand{\arraystretch}{1.2}
\setlength{\tabcolsep}{6pt}
\begin{tabular}{llll}
\toprule
\textbf{DSM-5 Category} & \textbf{Keyword\_KR} & \textbf{Keyword\_ENG} & \textbf{Count} \\
\midrule
A. Depressed mood & 실망 & Disappointment & 3,722 \\
A. Depressed mood & 우울 & Depression & 2,927 \\
A. Depressed mood & 힘들 & Struggling & 13,694 \\
A. Depressed mood & 힘들다 & Difficult & 1,332 \\
B. Loss of interest/pleasure & 무시 & Neglect & 8,226 \\
B. Loss of interest/pleasure & 소외 & Alienation & 2,609 \\
B. Loss of interest/pleasure & 혼자 & Alone & 18,482 \\
D. Insomnia or hypersomnia & 걱정 & Worry & 16,636 \\
D. Insomnia or hypersomnia & 고민 & Concern & 3,966 \\
D. Insomnia or hypersomnia & 스트레스 & Stress & 1,798 \\
D. Insomnia or hypersomnia & 잠 & Nap & 2,357 \\
E. Psychomotor agitation or retardation & 불안 & Anxiety & 9,716 \\
E. Psychomotor agitation or retardation & 짜증 & Annoyance & 2,868 \\
F. Fatigue & 무능 & Inability & 2,332 \\
G. Inappropriate guilt & 무능 & Incompetence & 2,332 \\
G. Inappropriate guilt & 후회 & Regret & 1,451 \\
H. Decreased concentration & 걱정 & Worry & 16,636 \\
H. Decreased concentration & 고민 & Concern & 3,966 \\
H. Decreased concentration & 불편 & Discomfort & 1,959 \\
H. Decreased concentration & 집중 & Focus & 1,971 \\
\bottomrule
\end{tabular}
\setlength{\abovecaptionskip}{12pt}  
\setlength{\belowcaptionskip}{0pt}   
\caption{Frequencies of 20 Keywords Across DSM-5 Symptom Categories}
\label{tab:dsm-mapping}
\end{table}

\section{Examples of Synthetic Data in KoACD}
\label{appendixG}

In Table~\ref{tab:synthetic-examples}, we provide one synthetic example per cognitive distortion, totaling 10.
\newpage

\begin{table}[H]
\centering
\renewcommand{\arraystretch}{1.4}
\setlength{\tabcolsep}{6pt}

\footnotesize
\begin{tabular}{m{3.5cm} m{12cm}}
\toprule
\textbf{Cognitive Distortion Type (English / Korean)} &
\multicolumn{1}{c}{\textbf{Example (English / Korean)}} \\
\midrule
All-or-Nothing Thinking \newline (흑백사고) & 
``I'm really bad at studying, and my grades are at the bottom. I can't even think about college, and getting a job seems impossible too. It feels like I have no future, like I’ve completely failed.'' \newline
(공부를 전혀 못해서 성적이 바닥이에요. 대학은 엄두도 못 내겠지만, 취업도 어렵겠죠. 결국 제대로 된 미래가 없을 것 같아 완전히 실패한 것 같아요.) \\
\midrule
Overgeneralization \newline (과잉일반화) & 
``I feel anxious because it seems like my classmates avoid talking to me. One day, I felt so left out that I cried. There have been so many times when everyone gathered and left me out. Now, I’m scared of being alone.'' \newline
(수업 시 친구들이 나와의 이야기를 피하는 것 같아 불안해. 하루는 소외된 기분이 들어 울었어. 모두가 모여서 나를 제외하고 나선 적이 많아, 이제 혼자가 될까 두려워.) \\
\midrule
Mental Filtering \newline (부정적 편향) & 
``I got my math test results—80 out of 100. It's over. Not even an A, and the top spot in the school is out of reach. Everything’s ruined. My future’s looking dark. I won’t get into college. I won’t be able to do anything. Maybe I should just give up.'' \newline
(수학 시험 성적표를 받았다. 80점. 망했다. A등급은 커녕, 전교 1등은 물 건너갔다. 모든 게 끝장났다. 내 미래는 어둡다. 대학도 못 갈 거야. 아무것도 안 될 거야. 그냥 포기해야겠다.) \\
\midrule
Discounting the Positive \newline (긍정 축소화) & 
``I wasn’t good at studying in middle school, but this time I finally got a score in the 60s. Instead of being happy for me, my parents got mad and said, ‘Is that something to brag about?’ It really hurt because it felt like all my effort didn’t matter.'' \newline
(내가 중학교 때는 공부를 잘 못했었는데, 이번에 겨우 60점대 맞았다고 자랑이냐며 부모님께서 화내셨어요. 노력한 게 인정받지 못하는 것 같아 너무 속상했습니다.) \\
\midrule
Jumping to Conclusions \newline (성급한 판단) & 
``There's a girl I like at my academy. I want to talk to her, but I'm scared I might get rejected or even end up being an outcast. If I confess and she’s not interested, it’ll hurt even more, so I keep hesitating. In the end, I’m just doing nothing.'' \newline
(학원에서 좋아하는 여자애가 있어요. 말을 걸고 싶지만, 왕따 당할까봐 두려워요. 고백하면 만약 저에게 관심이 없다면 더 힘들 것 같아서 늘 망설여져요. 결국 아무것도 못하고 있네요.) \\
\midrule
Magnification and Minimization \newline (확대와 축소) & 
``I’m having a hard time adjusting to my new school after transferring. I don’t have any friends, and keeping up with the lessons is tough, so I’m getting more and more depressed. I’m worried that if I stay isolated like this, I’ll never be happy.'' \newline
(전학 와서 새로운 학교에 잘 적응하지 못하고 있어요. 친구도 없고 공부도 따라가기 힘들어서 점점 우울해지고 있어요. 이렇게 고립되면 평생 행복할 수 없을 것 같아 걱정돼요.) \\
\midrule
Emotional Reasoning \newline (감정적 추론) & 
``I had an argument with my mom. I don’t think I can take this conflict anymore. If I go to school, I feel like I’ll just keep crying, but if I don’t go, there’ll be an even bigger fight. I honestly don’t know what to do.'' \newline
(엄마와 싸웠어요. 더 이상 갈등을 견딜 수 없을 것 같아요. 학교에 가면 계속 울고 있을 것 같고, 학교에 가지 않으면 더 큰 싸움이 벌어질 거예요. 과연 어떻게 해야 할지 모르겠어요.) \\
\midrule
``Should'' Statements \newline (``해야한다'' 진술) & 
``I know I should study hard for this exam, but it feels so tough every time I try, and I just want to give up. But I know I can’t, I have to keep going and work hard to get good grades.'' \newline
(이번 시험 준비를 잘해야 할 텐데, 공부할 때마다 너무 힘들어서 포기하고 싶어진다. 하지만 이렇게 해서는 안 되고, 반드시 열심히 공부해서 좋은 성적을 받아야 한다.) \\
\midrule
Labeling \newline (낙인찍기) & 
``I'm probably a loser because my test scores are bad. My friends will avoid me, and I’ll end up being a loner in high school too. I’m so clueless that I won’t be able to make any friends. I have no idea how I’m supposed to keep going.'' \newline
(시험 성적이 좋지 않은 내가 찐따일 거야. 친구들도 나를 피할 거고, 고등학교에서도 외톨이가 될 것 같아. 눈치 없는 나는 친구를 사귈 수 없을 거야. 앞으로 어떻게 살아갈지 막막하다.) \\
\midrule
Personalization \newline (개인화) & 
``I feel like my friends don’t like me. I joined a new club, but they’re leaving me out. I don’t even know what I did wrong. Even if I made a mistake, they shouldn’t treat me like this.'' \newline
(나는 친구들이 나를 싫어하는 것 같아. 새로운 동아리에 들어갔는데, 친구들이 나를 배제하고 있어. 내가 뭘 잘못했는지 모르겠어. 설령 내가 실수했더라도 이렇게 대할 순 없잖아.) \\
\bottomrule
\end{tabular}
\setlength{\abovecaptionskip}{12pt}  
\setlength{\belowcaptionskip}{0pt}   
\caption{Examples of Synthetic Data for Each Cognitive Distortion in KoACD}
\label{tab:synthetic-examples}
\normalsize  
\end{table}
\newpage
\section{Prompt Templates}
\label{appendixH}

We present the prompt templates used throughout the study for various stages of cognitive distortion identification, synthetic data generation, and evaluation. These prompts were designed to ensure consistency and accuracy across different processes.

To maintain conciseness, we replaced detailed descriptions and examples of cognitive distortions with the phrase \textit{``Refer to Table~\ref{tab:cognitive_distortions} for a detailed explanation of each cognitive distortion.''} 

This appendix includes Tables~\ref{tab:prompt-template-17}--\ref{tab:prompt-template-24}, which provide the full prompt templates for each stage.
\newpage
\begin{table}[H]
\centering
\footnotesize   
\renewcommand{\arraystretch}{1.0}
\setlength{\tabcolsep}{6pt}
\begin{tabular}{p{16cm}}
\toprule
\textbf{Analyzer} \\
\midrule
You are a psychology expert. \newline
Analyze the text below and, if a relevant cognitive distortion is present, select the most appropriate one. \newline
Choose from the following ten cognitive distortions: All-or-Nothing Thinking, Overgeneralization, Mental Filter, Discounting the Positive, Jumping to Conclusions, Magnification and Minimization, Emotional Reasoning, Should Statements, Labeling, and Personalization. \newline

\{previous\_cognitive\_distortions\} were deemed inappropriate in the previous analysis. Do not select them again under any circumstances. \newline

Previously rejected cognitive distortions: \{previous\_cognitive\_distortions\} \newline
Reason for rejection: \{previous\_reasons\} \newline

Since \{previous\_cognitive\_distortions\} were already deemed inappropriate: \newline
1. Do not select any of the above cognitive distortions again. \newline
2. You must choose only from the remaining cognitive distortions. \newline
3. If none of the remaining cognitive distortions are appropriate, respond with ``Unknown.'' \newline

When identifying cognitive distortions, carefully refer to the definitions and examples of the ten distortions to consider a variety of cognitive distortions. \newline
When deciding on a cognitive distortion, analyze the overall context of the text rather than focusing on a single sentence. \newline

Criteria for Responding with ``Unknown'': \newline
- The response requires speculation or subjective interpretation. \newline
- The intent of the sentence is unclear. \newline
- The speaker is not explicitly identified. \newline
- The text consists only of simple emotional expressions. \newline
- The text is merely a description of a situation or a question. \newline
- Context from prior conversations is necessary for understanding. \newline
- The text lacks value judgments or personal interpretation. \newline
- The meaning is unclear without external context. \newline
- The experience is described from another person's perspective. \newline
- Negative emotions are present, but no specific cognitive distortion is identifiable. \newline
- The text is a request for information, advice, or help. \newline
Important: If you determine ``Unknown,'' this is a final decision, and no further analysis or reconsideration is needed. If any of the above criteria apply, immediately respond with ``Unknown'' without considering alternative interpretations. \newline

Text to Analyze: \newline
\{input\_text\} \newline

List of Cognitive Distortions: \newline
Refer to Table 2 for a detailed explanation of each cognitive distortion. \newline

Analysis Request: \newline
1. When determining cognitive distortions, consider the overall context. \newline
2. Copy and paste all relevant sentences or paragraphs that support the selected cognitive distortion. Include at least two complete sentences. \newline
3. Provide a clear explanation for selecting the sentences, ensuring a logical cause-and-effect relationship in your reasoning. \newline
4. If no cognitive distortion applies, respond with ``Unknown.'' \newline

Output Format: \newline
- Cognitive Distortion: [Selected Cognitive Distortion] \newline
- Relevant Sentences/Paragraphs: [Text] \newline
- Reason for Selection: [Explanation] \newline

Additional Output Rules: \newline
- All responses must be grammatically complete sentences. \newline
- Sentences should not be cut off mid-thought. \newline
- The final sentence of the response must be fully structured and complete. \newline
- Do not use Markdown formatting. \newline
- When outputting [Selected Cognitive Distortion], do not select any distortions from \{previous\_cognitive\_distortions\}. \\ 
\bottomrule
\end{tabular}
\setlength{\abovecaptionskip}{12pt}  
\setlength{\belowcaptionskip}{0pt}   
\caption{Prompt for the analyzer role in the negotiation process}
\label{tab:prompt-template-17}
\normalsize
\end{table}

\begin{table}[H]
\centering
\renewcommand{\arraystretch}{1.0}
\setlength{\tabcolsep}{6pt}
\begin{tabular}{p{16cm}}
\toprule
\textbf{Evaluator} \\
\midrule
You are a psychology expert. \newline
Strictly evaluate the following cognitive distortion analysis provided by the analyzer. \newline
Refer to the cognitive distortions list for definitions and examples. \newline

Original Text: \newline
\{input\_text\} \newline

Analyzer’s Assessment: \newline
Cognitive Distortion: \{cognitive\_distortions\} \newline
Relevant Sentences/Paragraphs: \{related\_text\} \newline
Reason for Selection: \{reason\_text\} \newline

List of Cognitive Distortions: \newline
Refer to Table 2 for a detailed explanation of each cognitive distortion. \newline

Evaluation Rules: \newline
1. Is the selected cognitive distortion present in the text? \newline
- Assess whether the identified cognitive distortion can be reasonably inferred from the original text. \newline
- Do not rely on isolated sentences; patterns must be found within the overall flow of the text. \newline
2. Do the selected relevant sentences and reasoning properly support the cognitive distortion? \newline
- Check whether the selected sentences accurately align with the definition and examples of the cognitive distortion. \newline
- Evaluate whether the explanation logically connects the chosen sentences to the cognitive distortion. \newline
- Ensure that the justification is not overly interpretative or speculative. \newline

Judgment Criteria: \newline
- If any of the evaluation rules are violated, classify the analysis as ``Inappropriate.'' \newline
- If deemed inappropriate, clearly specify which rule was violated. \newline
- If the response is ``Unknown,'' accept it immediately. \newline

Output Format: \newline
Evaluation Result: [Appropriate / Inappropriate] \newline
Evaluation Reason: [Detailed explanation for each rule] \newline
Conclusion: \newline
[If appropriate] ``The current analysis is valid.'' \newline
[If inappropriate] ``The cognitive distortion should be reassessed.'' \newline

Additional Output Rules: \newline
- The evaluation reason must be fully structured in grammatically complete sentences. \newline
- Sentences should not be cut off mid-thought. \newline
- The final sentence must be a fully completed statement. \newline
- Do not use Markdown formatting. \\
\bottomrule
\end{tabular}
\setlength{\abovecaptionskip}{12pt}  
\setlength{\belowcaptionskip}{0pt}   
\caption{Prompt for the evaluator role in the negotiation process}
\label{tab:prompt-template-18}
\normalsize   
\end{table}

\begin{table}[H]
\centering
\renewcommand{\arraystretch}{1.2}
\setlength{\tabcolsep}{6pt}
\begin{tabular}{p{16cm}}
\toprule
\textbf{Independent Evaluator} \\
\midrule
You are a psychology expert. \newline
Thoroughly evaluate the appropriateness of the extracted cognitive distortion and its associated sentences/paragraphs. \newline

Content to Evaluate: \newline
Selected Cognitive Distortion: \{selected\_cognitive\_distortion\} \newline
Relevant Sentences: \{related\_sentences\} \newline

Evaluation Criteria (1--3 points): \newline
1 Point: Inappropriate \newline
- The relevant sentences do not contain the identified cognitive distortion. \newline
- OR the sentences are incomplete or lack clear context. \newline

2 Points: Partially Appropriate \newline
- The relevant sentences contain a cognitive distortion, but it does not match the selected one. \newline
- OR another cognitive distortion would be a better fit. \newline

3 Points: Appropriate \newline
- The relevant sentences clearly demonstrate the selected cognitive distortion. \newline
- The content aligns well with the definition and examples of the cognitive distortion. \newline

Output Format: \newline
Score: [1--3 points] \newline

Important Notes: \newline
- Only scores of 1, 2, or 3 may be used. \newline
- Intermediate scores (e.g., 1.5 or 2.5) are not allowed. \newline
- The evaluation rationale must be consistent with the assigned score. \\
\bottomrule
\end{tabular}
\setlength{\abovecaptionskip}{12pt}  
\setlength{\belowcaptionskip}{0pt}   
\caption{Prompt for independently verifying cognitive distortion}
\label{tab:prompt-template-19}
\normalsize   
\end{table}

\begin{table}[H]
\centering
\renewcommand{\arraystretch}{1.2}
\setlength{\tabcolsep}{6pt}
\begin{tabular}{p{16cm}}
\toprule
\textbf{Cognitive Clarification Method} \\
\midrule
Generate a realistic fictional adolescent story based on the given cognitive distortion and reference case. \newline
When writing the fictional story, ensure that the age and content remain within the adolescent range. \newline
Consider a variety of situations that may occur both inside and outside of school. \newline
Strictly follow the output format specified below. \newline

Input Information: \newline
Cognitive Distortion: \{cognitive\_distortions\} \newline
Relevant Real-Life Sentence/Paragraph: \{related\_sentences\} \newline
Original Text: \{input\_text\} \newline

Story Writing Requirements: \newline
1. Length: Must be 40 words or fewer (Exceeding 40 words is strictly prohibited). \newline
2. Format: [Gender/Age] --- [Story Content] \newline
3. Age: Must be between 13 and 19 years old. \newline
If gender, age, or school grade is mentioned in the original text, use that information to generate [Gender/Age]. \newline
(Gender: Male or Female, Middle School: 14--16 years old, High School: 17--19 years old) \newline
4. Theme: Events that occur in school, home, friendships, or daily adolescent life. \newline
5. Perspective: Write from a first-person point of view. \newline
6. Content: \newline
Clearly establish the situation (when, where, what, how). \newline
Maintain a logical cause-and-effect relationship within the story. \newline
The narrator (first-person) should naturally exhibit cognitive distortion. \newline

Constraints: \newline
1. The story must be inspired by the given real-life sentence, adapting it to a similar but new context. \newline
2. Utilize grammatical transformations, such as active/passive voice changes and word order modifications. \newline
3. Avoid starting the story with any of the following words:\{used\_words\} \newline
4. The word ``today'' must not be used. \newline
5. Do not explicitly mention cognitive distortion terms in the story. \newline
(e.g., Do NOT use terms like ``overgeneralization'' or ``all-or-nothing thinking.'') \newline

Output Format: \newline
[Gender/Age] --- [Generated Story] \newline

Important Notes: \newline
Ensure that the cognitive distortion characteristics reflected in the reference sentence are incorporated into the new story in a different yet relevant context. \newline
Do NOT exceed 40 words in the generated story (Strict limit: 40 words maximum). \\
\bottomrule
\end{tabular}
\setlength{\abovecaptionskip}{12pt}  
\setlength{\belowcaptionskip}{0pt}   
\caption{Prompt for cognitive clarification-based synthetic story generation}
\label{tab:prompt-template-20}
\normalsize   
\end{table}

\begin{table}[H]
\centering
\renewcommand{\arraystretch}{1.2}
\setlength{\tabcolsep}{6pt}
\begin{tabular}{p{16cm}}
\toprule
\textbf{Cognitive Balancing Method} \\
\midrule
Generate a realistic fictional adolescent story that reflects the characteristics of \{cognitive\_distortions\}, based on the provided real-life example. \newline
Strictly follow the output format specified below. \newline

Input Information: \newline
Original Text: \{input\_text\} \newline

Story Writing Requirements: \newline
1. Length: Must be 40 words or fewer (Exceeding 40 words is strictly prohibited). \newline
2. Format: [Gender/Age] --- [Story Content] \newline
3. Age: Must be between 13 and 19 years old. \newline
If gender, age, or school grade is mentioned in the original text, use that information to generate [Gender/Age]. \newline
(Gender: Male or Female, Middle School: 14--16 years old, High School: 17--19 years old) \newline
4. Theme: Events that occur in school, home, friendships, or daily adolescent life. \newline
5. Perspective: Write from a first-person point of view. \newline
6. Content: \newline
Clearly establish the situation (when, where, what, how). \newline
Maintain a logical cause-and-effect relationship within the story. \newline
The narrator (first-person) should naturally exhibit cognitive distortion. \newline

List of Cognitive Distortions: \newline
Refer to Table 2 for a detailed explanation of each cognitive distortion. \newline

Current Cognitive Distortion for Story Generation: \newline
\{cognitive\_distortions\} \newline

Constraints: \newline
1. The story must be inspired by the given real-life sentence, adapting it to a similar but new context. \newline
2. Utilize grammatical transformations, such as active/passive voice changes and word order modifications. \newline
3. Avoid starting the story with any of the following words:\{used\_words\} \newline
4. The word ``today'' must not be used. \newline
5. Do not explicitly mention cognitive distortion terms in the story. \newline
(e.g., Do NOT use terms like ``overgeneralization'' or ``all-or-nothing thinking.'') \newline

Output Format: \newline
[Gender/Age] --- [\{cognitive\_distortions\} Reflected Story] \newline

Important Notes: \newline
Do NOT exceed 40 words in the generated story (Strict limit: 40 words maximum). \\
\bottomrule
\end{tabular}
\setlength{\abovecaptionskip}{12pt}  
\setlength{\belowcaptionskip}{0pt}   
\caption{Prompt for cognitive balancing-based synthetic story generation}
\label{tab:prompt-template-21}
\normalsize   
\end{table}

\begin{table}[H]
\centering
\renewcommand{\arraystretch}{1.2}
\setlength{\tabcolsep}{6pt}
\begin{tabular}{p{16cm}}
\toprule
\textbf{Consistency Evaluator} \\
\midrule
Please evaluate the consistency of the given text and assign a single score between 1 and 3. \newline

Input Information: \newline
Selected Cognitive Distortion: \{selected\_cognitive\_distortion\} \newline
Relevant Sentences: \{related\_sentences\} \newline
Generated Story: \{generated\_story\} \newline

Evaluation Criteria (1--3 points): \newline
Assess whether the selected cognitive distortion is neither exaggerated nor minimized and whether the original meaning of the relevant sentences is preserved while being appropriately expressed in the generated story. \newline
- Is the selected cognitive distortion accurately maintained without distortion from the relevant sentences? \newline
- Has the meaning of the relevant sentences been appropriately conveyed in the generated story without excessive modification? \newline
- Does the generated story logically align with the selected cognitive distortion and its context? \newline

Scoring Guidelines: \newline
1 Point: The selected cognitive distortion or the context of the relevant sentences is significantly distorted or altered in the generated story. \newline
2 Points: The selected cognitive distortion and the context of the relevant sentences are partially retained, but there are some inconsistencies or unnatural expressions. \newline
3 Points: The selected cognitive distortion and the context of the relevant sentences are naturally maintained, forming a logically coherent story. \newline

Output Format: \newline
Score: [1--3 points] \newline

Important Notes: \newline
- Only scores of 1, 2, or 3 may be used. \newline
- Intermediate scores (e.g., 1.5 or 2.5) are not allowed. \newline
- The evaluation rationale must be consistent with the assigned score. \\
\bottomrule
\end{tabular}
\setlength{\abovecaptionskip}{12pt}  
\setlength{\belowcaptionskip}{0pt}   
\caption{Prompt for consistency evaluation of synthetic data}
\label{tab:prompt-template-22}
\normalsize   
\end{table}

\begin{table}[H]
\centering
\renewcommand{\arraystretch}{1.2}
\setlength{\tabcolsep}{6pt}
\begin{tabular}{p{16cm}}
\toprule
\textbf{Accuracy Evaluator} \\
\midrule
Please evaluate the accuracy of the given text and assign a single score between 1 and 3. \newline

Input Information: \newline
Selected Cognitive Distortion: \{selected\_cognitive\_distortion\} \newline
Relevant Sentences: \{related\_sentences\} \newline
Generated Story: \{generated\_story\} \newline

Evaluation Criteria (1--3 points): \newline
Evaluate whether the generated story is correctly classified under the most relevant cognitive distortion among the ten defined categories. \newline
- Does the selected cognitive distortion correctly classify the cognitive distortion present in both the relevant sentences and the generated story? \newline
- When compared to other cognitive distortions, is the selected cognitive distortion the most appropriate choice? \newline
- Is there a logical consistency between the selected cognitive distortion and the way it is expressed in the generated story? \newline

Scoring Guidelines: \newline
1 Point: The selected cognitive distortion significantly mismatches the cognitive distortion found in the relevant sentences and the generated story. \newline
2 Points: The selected cognitive distortion is partially appropriate, but another cognitive distortion might be a better fit. \newline
3 Points: The selected cognitive distortion is the most accurate classification of the cognitive distortion found in the relevant sentences and the generated story. \newline

Output Format: \newline
Score: [1--3 points] \newline

Important Notes: \newline
- Only scores of 1, 2, or 3 may be used. \newline
- Intermediate scores (e.g., 1.5 or 2.5) are not allowed. \newline
- The evaluation rationale must be consistent with the assigned score. \\
\bottomrule
\end{tabular}
\setlength{\abovecaptionskip}{12pt}  
\setlength{\belowcaptionskip}{0pt}   
\caption{Prompt for accuracy evaluation of synthetic data}
\label{tab:prompt-template-23}
\normalsize   
\end{table}

\begin{table}[H]
\centering
\renewcommand{\arraystretch}{1.2}
\setlength{\tabcolsep}{6pt}
\begin{tabular}{p{16cm}}
\toprule
\textbf{Fluency Evaluator} \\
\midrule
Please evaluate the fluency of the given text and assign a single score between 1 and 3. \newline

Input Information: \newline
Generated Story: \{generated\_story\} \newline

Evaluation Criteria (1--3 points): \newline
Evaluate whether the generated story is grammatically sound and maintains human-like fluency in its structure and readability. \newline
- Is the sentence structure natural and fluent? \newline
- Are there any grammatical errors? \newline
- Is the flow between sentences smooth, making the overall story cohesive? \newline

Scoring Guidelines: \newline
1 Point: The text contains many grammatical errors or is highly unnatural. \newline
2 Points: The text has minor grammatical issues or slightly awkward expressions but is still generally understandable. \newline
3 Points: The text is grammatically correct and reads naturally with a smooth sentence structure. \newline

Output Format: \newline
Score: [1--3 points] \newline

Important Notes: \newline
- Only scores of 1, 2, or 3 may be used. \newline
- Intermediate scores (e.g., 1.5 or 2.5) are not allowed. \newline
- The evaluation rationale must be consistent with the assigned score. \\
\bottomrule
\end{tabular}
\setlength{\abovecaptionskip}{12pt}  
\setlength{\belowcaptionskip}{0pt}   
\caption{Prompt for fluency evaluation of synthetic data}
\label{tab:prompt-template-24}
\normalsize   
\end{table}

\clearpage

\section{Evaluation Form}
\label{appendixI}
\begin{table}[H]
\centering

\renewcommand{\arraystretch}{1.2}
\setlength{\tabcolsep}{6pt}
\begin{tabular}{p{16cm}}
\toprule
\textbf{Expert Evaluation Form (Korean)} \\
\midrule
안녕하세요. \newline
\newline
소중한 시간 내어 주셔서 진심으로 감사드립니다. \newline

본 작업은 청소년의 실제 발화 데이터를 바탕으로, 각 문장에 드러난 인지 왜곡의 유형과 문장 자체의 표현 완성도를 함께 평가해주시는 것입니다. \newline
본 평가에서 수집되는 모든 결과는 연구 목적과 데이터 해석에만 사용되며, 상업적 용도나 연구 외 배포는 이루어지지 않습니다. \newline
\newline
평가에 필요한 문장들은 별도의 파일로 제공되며, 각 문장에 대해 세 가지 항목을 중심으로 판단해주시면 됩니다. \newline
\newline

첫 번째는 선택된 인지 왜곡이 과장되거나 축소되지 않고, 관련 문장의 본래 의미를 유지하면서 생성된 이야기에 적절하게 표현되었는지를 종합적으로 평가하는 것입니다. \newline
표시된 인지 왜곡 유형이 그 문장을 잘 설명하고 있는지를 1점에서 3점 사이로 평가해 주세요. \newline

선택 기준은 다음과 같습니다: \newline
1점: 선택된 인지 왜곡이나 관련 문장의 상황이 생성된 이야기에서 크게 왜곡되거나 변형됨 \newline
2점: 선택된 인지 왜곡과 관련 문장의 상황이 생성된 이야기에 부분적으로 유지되었으나, 일부 불일치하거나 부자연스러움 \newline
3점: 선택된 인지 왜곡과 관련 문장의 상황이 생성된 이야기에 자연스럽게 유지되며, 논리적으로 일관된 이야기를 구성함 \newline
\newline

두 번째는 문장의 정확성에 대한 평가입니다. \newline
생성된 이야기가 10가지 인지 왜곡 중 가장 적절한 항목으로 정확하게 반영되었는지 평가합니다. \newline

이 항목 역시 1점부터 3점까지 평가하며, 기준은 다음과 같습니다: \newline
1점: 선택된 인지 왜곡이 관련 문장과 생성된 이야기의 인지 왜곡 유형과 크게 불일치함 \newline
2점: 선택된 인지 왜곡이 부분적으로 적절하나, 다른 인지 왜곡이 더 적합할 수 있음 \newline
3점: 선택된 인지 왜곡이 관련 문장과 생성된 이야기의 인지 왜곡을 가장 정확하게 분류함 \newline
\newline

세 번째는 문장의 유창성에 대한 평가입니다. \newline
문장이 문법적으로 자연스럽고, 읽기 쉬우며, 의미 흐름이 잘 이어지는지를 평가해 주세요. \newline
특히, 문장 안에서 원인–문제–해결 간의 논리적 연결 구조가 유지되고 있는지도 함께 고려해 주시면 됩니다. \newline

이 항목 역시 1점부터 3점까지 평가하며, 기준은 다음과 같습니다: \newline
1점: 문장이 부자연스럽거나 문법적 오류가 많아 이해하기 어려움 \newline
2점: 다소 어색하거나 일부 논리적 연결이 부족하지만 대체로 의미 전달 가능함 \newline
3점: 문장이 매끄럽고 자연스러우며, 구조적으로도 논리적 일관성이 잘 유지됨 \newline

아울러, 인지 왜곡 분류와 관련해 문장이 다른 유형으로도 해석될 수 있다고 판단되실 경우, \newline
그 가능성과 간단한 이유를 덧붙여주시면 감사하겠습니다. \newline

작업 중 부담이 느껴지시거나 일정 조정이 필요하신 경우, 언제든지 중단 또는 조율 가능하니 편히 말씀해 주세요. \newline
\newline
문의 사항이 있으실 경우에도 언제든 연락 주시기 바랍니다. \newline
\newline
다시 한 번 깊이 감사드립니다. \\
\bottomrule
\end{tabular}
\setlength{\abovecaptionskip}{12pt}  
\setlength{\belowcaptionskip}{0pt}   
\caption{Expert Evaluation Form for Cognitive Distortion (Korean)}
\label{tab:expert-evaluation-form}
\normalsize   
\end{table}

\begin{table}[H]
\centering
\renewcommand{\arraystretch}{1.2}
\setlength{\tabcolsep}{6pt}
\begin{tabular}{p{16cm}}
\toprule
\textbf{Expert Evaluation Form (English)} \\
\midrule
Hello, \newline
Thank you very much for taking the time to participate in this task. \newline

This task involves evaluating each sentence from real adolescent utterances based on both the type of cognitive distortion it contains and the overall quality of the sentence’s expression. \newline
All results collected in this evaluation will be used solely for research purposes and data analysis, and will not be used for commercial purposes or distributed outside the research context. \newline
The sentences to be evaluated are provided in a separate file, and for each sentence, please make your assessment according to the following three criteria. \newline

\textbf{Consistency:} \newline
Please evaluate whether the selected cognitive distortion is neither exaggerated nor minimized and whether it is appropriately represented in the generated story while preserving the original meaning of the related sentence. Assess how well the indicated cognitive distortion type explains the sentence, using a 1-to-3 point scale. \newline
1 point: The selected cognitive distortion or the situation in the related sentence is significantly distorted or altered in the generated story. \newline
2 points: The selected cognitive distortion and the situation in the related sentence are partially maintained in the generated story, but there are some inconsistencies or unnatural expressions. \newline
3 points: The selected cognitive distortion and the situation in the related sentence are naturally preserved in the generated story, forming a logically consistent narrative. \newline

\textbf{Accuracy:} \newline
Please evaluate whether the generated story correctly reflects the most appropriate category among the 10 cognitive distortion types. \newline
1 point: The selected cognitive distortion significantly disagrees with the type present in the related sentence and generated story. \newline
2 points: The selected cognitive distortion is partially appropriate, but another distortion type might be more suitable. \newline
3 points: The selected cognitive distortion most accurately matches the cognitive distortion in the related sentence and generated story. \newline

\textbf{Fluency:} \newline
Please assess whether the sentence is grammatically natural, easy to read, and coherent in terms of meaning. In particular, consider whether the cause–problem–solution structure within the sentence is logically connected. \newline
1 point: The sentence is unnatural or contains many grammatical errors, making it difficult to understand. \newline
2 points: The sentence is somewhat awkward or partially lacks logical flow, but overall meaning can be conveyed. \newline
3 points: The sentence is smooth, natural, and maintains structural and logical consistency. \newline

In addition, if you believe that a sentence could be interpreted as a different type of cognitive distortion, please indicate the possibility and provide a brief explanation. \newline

If you feel any burden during the task or need to adjust your schedule, please feel free to pause or coordinate as needed. \newline
If you have any questions, do not hesitate to contact us. \newline
Once again, thank you very much for your contribution. \\
\bottomrule
\end{tabular}
\setlength{\abovecaptionskip}{12pt}  
\setlength{\belowcaptionskip}{0pt}   
\caption{Expert Evaluation Form for Cognitive Distortion (English)}
\label{tab:eval-form-english}
\normalsize
\end{table}

\end{document}